\documentclass{article}
\usepackage{spconf,amsmath,graphicx}
\usepackage{multirow, subcaption}
\usepackage{hyperref}
\usepackage{comment}

\newcommand{\rottext}[2]{\parbox[t]{2mm}{\multirow{#1}{*}{\rotatebox[origin=c]{90}{#2}}}
}
\usepackage{color}

\title{Prosodic Alignment for off-screen automatic dubbing}
\name{Yogesh Virkar,  Marcello Federico, Robert Enyedi, Roberto Barra-Chicote}
\address{Amazon AI}

\begin{document}
\ninept

\maketitle

\begin{abstract}
The goal of automatic dubbing is to perform speech-to-speech translation while achieving audiovisual coherence. This entails {\em isochrony}, i.e., translating the original speech by also matching its prosodic structure into phrases and pauses, especially when the speaker's mouth is visible. In previous work, we introduced a {\em prosodic alignment} model to address isochrone or {\em on-screen} dubbing. In this work, we extend the prosodic alignment model to also address {\em off-screen} dubbing that requires less stringent synchronization constraints. We conduct experiments on four dubbing directions – English to French, Italian, German and Spanish – on a publicly available collection of TED Talks and on publicly available YouTube videos. Empirical results show that compared to our previous work the extended prosodic alignment model provides significantly better subjective viewing experience on videos in which on-screen and off-screen automatic dubbing is applied for sentences with speakers mouth visible and not visible, respectively. 
\end{abstract}

\begin{keywords}
speech translation, text-to-speech, automatic dubbing, off-screen dubbing
\end{keywords}

\section{Introduction}
\label{sec:intro}

Automatic Dubbing (AD) is an extension of speech-to-speech translation that replaces speech in a video with speech in a different language while preserving as much as possible the user experience. Speech translation \cite{casacuberta_recent_2008, weiss_sequence--sequence_2017, cross_vila_end--end_2018, sperber_speech_2020} consists of recognizing a speech utterance in the source language, performing translation, and optionally resynthesizing speech in the target language. Use cases for speech translation include human-to-human interaction, live lectures, etc. in which close to real-time response is needed. In contrast, AD is used to automate the localization of audiovisual content, a highly complex workflow \cite{chaume_synchronization_2004} usually managed during post-production by dubbing studios. High quality video dubbing usually involves speech synchronization at the utterance level (isochrony), lip movement level (phonetic synchrony) and body movement level (kinetic synchrony). In the past, most work on AD \cite{oktem_prosodic_2019, federico_speech--speech_2020, federico_evaluating_2020, virkar_improvements_2021} addressed isochrony, i.e., translating original speech by optimally matching its sequence of phrases and pauses. The idea is to first machine translate the source transcript by generating output with roughly the same duration \cite{saboo_integration_2019, lakew_controlling_2019} --i.e. in terms of number of characters or syllables -- of the input. Next, the translation is segmented into phrases and pauses 
of the same duration as that of the original phrases. This step is called prosodic alignment (PA). 

Past work on PA \cite{oktem_prosodic_2019, federico_speech--speech_2020, federico_evaluating_2020, virkar_improvements_2021} focused on isochrony in the context of on-screen dubbing, i.e., dubbing of videos in which the speaker's mouth is visible for all utterances. However, in practical settings, it is quite common that videos contain scenes in which the speaker is not visible (off-screen) and for which the synchronization constraints of isochrone dubbing can be relaxed. To address this case, we extend PA with a mechanism to address on/off-screen dubbing in which all or some of the sentences in a video are off-screen. We perform automatic and human evaluations that compare our original PA model for isochrone dubbing \cite{virkar_improvements_2021} with the augmented PA model for on/off dubbing \footnote{\textit{We will provide a link to examples of dubbed videos with the camera ready version of the paper.}}. We report results on a test set of TED talks extracted from the MUST-C corpus \cite{MUSTC_2019} and on 3 publicly available YouTube videos, on four dubbing directions, English (en) to French (fr), Italian (it), German (de) and Spanish (es). To summarize, our contributions in this work are: 
\begin{itemize}
    \item We extend the PA model \cite{virkar_improvements_2021} to address off-screen dubbing. 
    \item We introduce an automatic metric to compute intelligibility of dubbed videos. 
    \item We run extensive automatic and subjective human evaluations comparing previous work with the new PA model on TED Talks and YouTube clips.
    \item Finally, we provide extensive analysis using the linear mixed-effects models to demonstrate the utility of automatic metrics in predicting human score. 
    %that the linear combination of the automatic metrics can predict scores that are well correlated with the average human score of dubbed video viewing experience.
\end{itemize}

Our paper is organized as follows: First, we describe our dubbing architecture, then, we focus on existing and new PA methods and finally present and discuss experimental results comparing past and current work. 

\section{Dubbing Architecture}
\begin{figure} [h]
    \centering
    \includegraphics[scale=0.4]{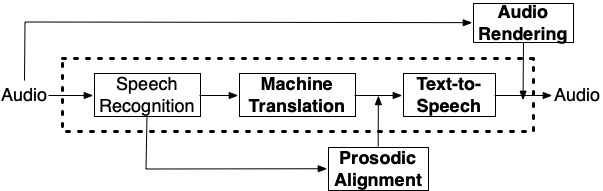}
    \caption{Speech translation pipeline (dotted box) with enhancements introduced to perform automatic dubbing (bold).}
    \label{fig:dubarch}
\end{figure}

We build on the automatic dubbing architecture presented in \cite{federico_evaluating_2020, federico_speech--speech_2020}. Figure~\ref{fig:dubarch} shows (in bold) how we extend a speech-to-speech translation \cite{casacuberta_recent_2008, weiss_sequence--sequence_2017, cross_vila_end--end_2018} pipeline with: neural machine translation (MT) robust to ASR errors and able to control verbosity of the output \cite{lakew_controlling_2019, vaswani_attention_2017, di_gangi_robust_2019}; prosodic alignment (PA) \cite{oktem_prosodic_2019, federico_evaluating_2020, virkar_improvements_2021} which addresses phrase-level synchronization of the MT output by leveraging the force-aligned source transcript; neural text-to-speech (TTS) \cite{prateek_other_2019, latorre_effect_2018, lorenzo-trueba_towards_nodate} with precise duration control; and, finally, audio rendering that enriches TTS output with the original background noise (extracted via audio source separation with deep U-Nets \cite{ronneberger2015u, jansson_singing_2017}) and reverberation, estimated from the original audio \cite{lollmann2010improved, habets2006room}.

\section{Related Work}
In the past there has been little work to address isochrony in dubbing \cite{oktem_prosodic_2019, federico_speech--speech_2020, federico_evaluating_2020, virkar_improvements_2021}. The approach of \cite{oktem_prosodic_2019} involved generating and rescoring segmentation hypotheses by utilizing the attention weights of neural machine translation. While they focused only on the linguistic content matching between source-target phrases, Federico et al. \cite{federico_speech--speech_2020} focused on fluency. In particular, the model of \cite{federico_speech--speech_2020} utilized source-target duration matches and dynamic programming search for faster implementation. In their subsequent works \cite{federico_evaluating_2020, virkar_improvements_2021} they further enhanced the prosodic alignment model by addition of features controlling for speaking rate variation and linguistic content matching. Additionally, they introduced time-boundary relaxation to further improve speaking rate control. However, none of these works focused on relaxing isochrony constraints by considering if the speaker is on-screen or off-screen. Recently, \cite{karakanta_two_2020} leveraged  on/off screen information to improve MT of dubbing scripts. Their rationale is that as human translations of scripts used in training reflect the different sync requirements posed by on-screen and off-screen speech, it is worth introducing the same bias in the neural MT model. Our work complements \cite{karakanta_two_2020}, by showing how to leverage the same information in order to improve prosodic alignment, too. 

\section{Prosodic Alignment}
\begin{figure} [t]
    \centering
    \includegraphics[scale=0.16]{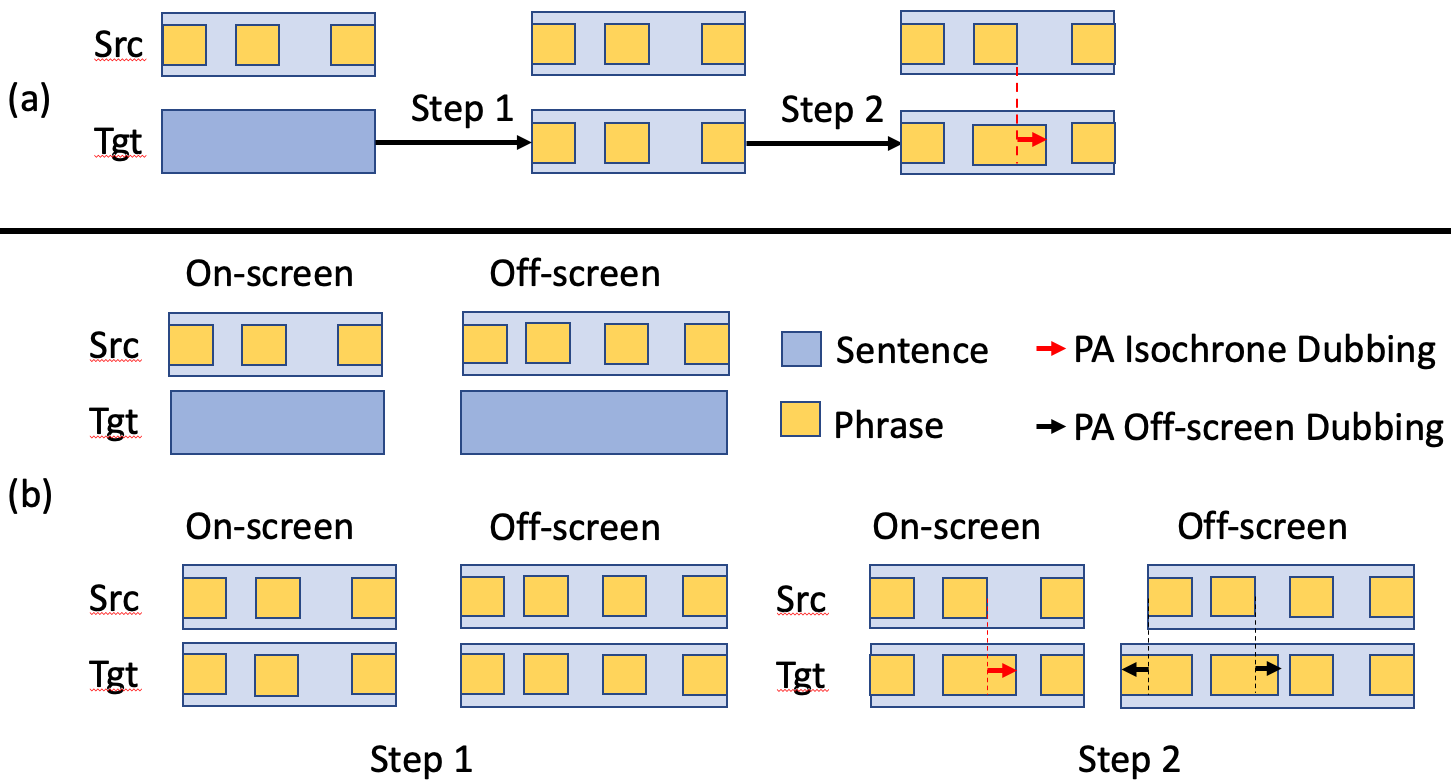}
    \caption{Overview of dubbing conditions: (a) Isochrone dubbing  \cite{virkar_improvements_2021} and (b) On/Off dubbing. The length of a box corresponds to the duration.}
    \label{fig:dub_cond_overview}
\end{figure}

\subsection{Isochrone Dubbing}

PA \cite{oktem_prosodic_2019, federico_speech--speech_2020, federico_evaluating_2020, virkar_improvements_2021} aims to segment a translation to optimally match the sequence of phrases and pauses of the corresponding source utterance. Let $\mathbf{e}$ denote the source sentence with $n$ words and $k$ breakpoints denoted by $\mathbf{i} = i_1, \dots, i_k$ such that $1 \leq i_1 \leq i_2 \leq \dots i_k = n$. Let $T$ denote the temporal duration of $\mathbf{e}$ and let $\mathbf{s}$ denote a temporal segmentation into $k$ segments where $\Delta\epsilon$ is the minimum silence after and before each break point.\footnote{In this work, the minimum silence interval $\Delta\epsilon$ is set to $300ms$.} Given the target sentence $\mathbf{f}$ of $m$ words, the goal of PA is to find $k$ breakpoints $\mathbf{j} = 1 \leq j_1 < j_2 < \dots j_k = m$  within $\mathbf{f}$ that  maximize the probability: 
\begin{align}
    \max_{\mathbf{j}} \log \Pr \left(\mathbf{j} | \mathbf{i}, \mathbf{e}, \mathbf{f}, \mathbf{s}\right).
\end{align}
\noindent
Assuming a Markovian model of $\mathbf{j}$, we get: \begin{align}
    \Pr \left(\mathbf{j} | \mathbf{i}, \mathbf{e}, \mathbf{f}, \mathbf{s}\right) = \displaystyle\sum_{t=1}^{k} \log\Pr\left(j_t | j_{t-1}; t, \mathbf{i}, \mathbf{e}, \mathbf{f}, \mathbf{s}\right) \enspace.
\end{align}
In \cite{federico_evaluating_2020} we derive a recurrent formular which permits to efficiently solve (2) with dynamic programming.  
Moreover, we allow target segments to possibly extend or contract the duration of the corresponding source interval by some fraction of $\Delta\epsilon$ to the left and to the right, denoted by $\delta_l$ and $\delta_r$ respectively, s.t., $\delta_l, \delta_r \in \left\{0, \pm\frac{1}{4}, \pm\frac{2}{4}, \pm\frac{3}{4}, \pm\frac{4}{4}\right\}$. In this way,  we  trade off strict isochrony for small adjustments to the speaking rate, in order to improve the viewing experience. In the past work \cite{virkar_improvements_2021} it was observed that relaxations on isochrony do not help improve the accuracy of finding optimal segmentation but only help improve speech fluency and smoothness. 

{\bf{Two-step optimization procedure:}} For the above reasons, in \cite{virkar_improvements_2021}, we introduce a two-step optimization procedure. 
In Step 1, we optimize the weights of the following log-linear model by maximizing segmentation accuracy over a manually annotated data set:  
\begin{align}
    \log \Pr\left(j|j^{'};t\right) \propto \displaystyle\sum_{a=1}^{4} w_a\log s_a\left(j, j^{'}; t\right) \label{eq:models1}
\end{align}
The feature functions $s_a$ of the model (notice that we dropped some of the dependencies in eq. (2) for readability) denote -- (1) the language model score of target break point, (2) the cross-lingual semantic match score across source and target segments, (3) the speaking rate variation across target segments and (4) the speaking rate match across source and target segments respectively.

In Step 2, starting from given breakpoints $\mathbf{\hat{j}}$,  we optimize the relaxations $\delta_l$ and $\delta_r$ for the $t$-th segment using another recurrent equation, that can also be solved via dynamic programming, derived from the log-linear model:  
\begin{align}
    \log \Pr\left(\delta_l, \delta_r|\dots;t\right) \propto \displaystyle\sum_{a=1}^{5} w_a\log s_a\left(\delta_l, \delta_r, \hat{j}_t, \hat{j}_{t-1},\delta^{'}_l, \delta^{'}_r; t\right) \label{eq:models2}
\end{align}
which includes as additional feature $s_5$  the isochrony score \cite{virkar_improvements_2021}. 
\noindent
Weight $w_5$ is optimized by maximizing speech smoothness \cite{virkar_improvements_2021} over the training set, assuming the reference breakpoints $\mathbf{\hat{j}}$ are given. Speech smoothness measures speaking rate variations across contiguous segments. Speaking rate computations rely on the strings $\tilde{f}_t$ and $\tilde{e}_t$, denoting the $t$-th source and target segments, as well as the original interval $s_t$ and the relaxed interval $s^{*}_t$. Hence, the speaking rate of a source
(target) segment is computed by taking the ratio between the duration of the utterance by source (target) TTS run at normal speed and the source (target) interval length, \footnote{We run TTS on the entire sentence, force-align audio with text \cite{ochshorn_gentle_2017, mcauliffe_montreal_2017} and compute segment duration from the time-stamps of the words.}, i.e:
\begin{align}
    r_e(t) &= \frac{\text{duration}(\text{TTS}_e(\tilde{e}_t))}{|s_t|}\\
    r_f(t) &= \frac{\text{duration}(\text{TTS}_f(\tilde{f}_t))}{|s^{*}_t|}
\end{align}

\subsection{On/Off Dubbing}
\label{subsec:relmixdub}
\noindent
Figure~\ref{fig:dub_cond_overview}(a) shows that in our past work \cite{virkar_improvements_2021}, during inference we apply the two steps of the PA component at the level of a single sentence, i.e., for each target sentence we first segment and then find the optimal relaxation using trained model defined by Eqs.~\eqref{eq:models1}, \eqref{eq:models2}.  In this work we focus on the more general use case of dubbing videos in which some of the sentences are on-screen -- i.e. the speaker's mouth is visible -- and some are off-screen -- i.e., the speaker's mouth is not visible. We name this general case on/off dubbing. For off-screen sentences, we do not need the stringent requirement of isochrony and we can perform more relaxation to further improve the speaking rate control. 
Figure~\ref{fig:dub_cond_overview}(b) gives an overview of the algorithm for on/off dubbing which extends the two-step 
isochrone dubbing algorithm as follows: 
\begin{itemize}
    \item[Step 1] We apply the segmentation step (Eq.~\eqref{eq:models1}) for all sentences, i.e., both on and off screen sentences. 
    \item[Step 2] We apply the relaxation step locally for all on-screen sentences using Eq.~\eqref{eq:models2} and globally across all off-screen sentences by replacing Eq.~\eqref{eq:models2} with  the following:

\begin{align}
    \Pr\left(\delta_l, \delta_r|\dots; t\right) \propto \begin{cases}
    1 & \text{if }r_f(t) \leq 1  \enspace,\\
    2 - r_f(t) & \text{if } 1 < r_f(t) \leq 2 \enspace,\\
    0 & \text{if }r_f(t) > 2
    \end{cases} \label{eq:globalrelax}
\end{align}
\end{itemize}

In Step 2 we also apply a more relaxed policy in allocating relaxations $\delta_l$ and $\delta_r$ inside off-screen sentences. In particular, we allow using the entire available inter-phrase and inter-sentence intervals rather than limiting them to maximum $\Delta\epsilon$. Isochrone dubbing utilizes relaxation mechanism locally inside each sentence since for on-screen sentences, we tradeoff isochrony for improved speaking rate control. In contrast, for off-screen sentences we do not need isochrony and hence we utilize a \textit{global relaxation mechanism} by computing optimal relaxations across all off-screen sentences using dynamic programming. 

Regarding the scoring function (\ref{eq:globalrelax}), when the target speaking rate $r_f(t)$ is below $1$, it returns a maximum score since for off-screen phrases we can contract time boundaries by setting the speaking rate $r_f(t)$ to $1$ without consequences. When the speaking rate of a target phrase $r_f(t)$ is larger than $2$, it returns the lowest score since too high speaking rates will result in unintelligible TTS speech. For $1 \leq r_f(t) \leq 2$, it returns scores that increasingly penalize values larger 
\noindent
than 1, as they will correspond to less and less intelligible TTS speech.

\section{Evaluation Data and Metrics}
\label{sec:evaluation_data}

For training and evaluation, we re-translated and annotated video clips from 20 TED talks of the MUST-C corpus \cite{di_gangi_must-c:_2019} and 3 YouTube videos by vloggers. Each video clip contains 4 sentences manually annotated for on or off screen \footnote{We consider the mixed case, in which the speaker's mouth is visible only for some part of the sentence, to be on screen so as to preserve isochrony.}. A single sentence contains one or more pauses of at least 300ms that are detected by force-aligning original English audio with text \cite{ochshorn_gentle_2017}. We manually collected and segmented translations in 4 languages - French, Italian, German and Spanish - using external vendors to fit duration and segmentation of corresponding English utterances.

Overall, we created two test sets to test on/off dubbing PA (ON/OFF) against Isochrone dubbing PA (ISO), (i) $D_1$: 15 4-sentence clips in which all clips have all sentences being off-screen, (ii) $D_2$: 15 4-sentence clips in which all clips have at least one sentence being on-screen. Compared to our previous work \cite{virkar_improvements_2021}, we increase the size of extracted clips from 1 sentence to 4 sentences to test if the global relaxation mechanism provides better subjective viewing experience. To automatically estimate quality of dubbing, similar to \cite{federico_evaluating_2020, virkar_improvements_2021} we define Fluency (F) and Smoothness (Sm). For Smoothness we consider contiguous segments that span an entire 4-sentence video clip as opposed to a single sentence. We additionally introduce the following metric:

\noindent
\textbf{Intelligibility} (In) of audio by dubbing target sentences $\mathbf{F}$ using prosodic alignment is defined by the ratio: 
\begin{align}
    I(\mathbf{F}) = \displaystyle\frac{1 - \mathrm{WER}(\mathrm{TTS_f}(\mathrm{PA}(\mathbf{F})))}{1 - \mathrm{WER}(\mathrm{TTS_f}(\mathbf{F}))}
\end{align}
where WER is the word error rate by an automatic speech recognition system \footnote{We use the off-the-shelf service Amazon Transcribe\\ (\href{https://aws.amazon.com/transcribe}{https://aws.amazon.com/transcribe}).} run on  TTS audio, either with prosody-alignment (numerator) or without prosody alignment (denominator).

\section{Experiments}

\subsection{Automatic Evaluation}
\begin{table}[t]
    \centering
    \setlength\tabcolsep{5pt}
    \begin{tabular}{l|l|l|ll|ll}
    & &    & \multicolumn{2}{c|}{$D_1$} &    \multicolumn{2}{c}{$D_2$}  \\ 
    & &    & \multicolumn{2}{l|}{ISO\hspace{0.1cm}      \hspace{0.0cm} ON/OFF} &    \multicolumn{2}{l}{ISO\hspace{0.1cm}      \hspace{0cm} ON/OFF} \\ 
      \hline
      %Wins  \hspace{0cm} & 28.37\% & 46.33\% ** & 29.9\% & 42.24\% ** \\
    \rottext{12}{MuST-C}
    & fr & Sm  & 68.5 & 75.3$^\circ$ & 60.7 & 69.3$^*$ \\
         && F      & 76.7  & 83.3 & 61.3 & 72.6 \\
         && In     & 93.6  & 93.5 & 93.5 & 93.2 \\
    & it & Sm  & 58.7  & 75.3$^*$ & 52.0 & 68.3$^*$ \\
         && F      & 68.3  & 80.0$^\circ$ & 54.0 & 68.3$^*$\\
         && In     & 117.2 & 121.1$^*$ & 98.8 & 99.0 \\
    & de & Sm & 66.4  & 79.7$^*$  & 57.6 & 70.4$^*$ \\
         && F      & 81.7  & 86.7 & 58.6 & 74.1$^*$  \\
         && In     & 94.3  & 94.3 & 91.7 & 93.0  \\
     & es & Sm & 71.6 & 82.0$^*$ & 61.9 & 76.0$^*$  \\
          && F      & 80.0 & 85.0  & 61.9 & 79.4$^*$ \\
          && In     & 124.9 & 125.7& 97.0 & 98.0 \\
    \hline
    \rottext{12}{YouTube}
    & fr & Sm  & 70.6 & 80.9$^*$ & 70.7 & 73.2 \\
         && F   & 66.7 & 80.0 & 60.0 & 60.0 \\
         && In   & 102.7 & 103.9 & 99.8 & 102.8 \\
         
    & it & Sm  & 73.6 & 81.3$^*$ & 66.1 & 64.7  \\
     && F   & 40.0 & 73.3&  46.7 & 43.8\\
     && In   & 109.5 & 111.3 & 101.4 & 101.5\\
     
     & de & Sm  & 69.9 & 82.3$^*$ & 61.9 & 67.1$^\circ$ \\
     && F   & 53.3 & 66.7 & 46.7 & 53.3\\
     && In   & 102.7 & 105.6$^*$ & 93.8 & 100.3$^\circ$\\
     
     & es & Sm  & 70.1 & 78.1$^*$ & 67.0 & 72.0$^*$\\
     && F   & 33.3 & 60.0 & 60.0 & 66.7\\
     && In   & 105.6 & 108.1 & 106.1 & 107.7$^\circ$\\
     
    \hline
    \end{tabular}
    \caption{Automatic evaluation of PA variants in terms of Smoothness (Sm), Fluency (F), Intelligibility (In) of: isochrone dubbing PA\cite{virkar_improvements_2021} (ISO),  mixed dubbing PA (ON/OFF) applied on off-screen clips ($D_1$) or on mixed off-screen and on-screen clips ($D_2$). All test sets consist of 15 4-sentence video clips for each domain (MuST-C, YouTube). Significance testing is done with levels $p<0.05$ ($^\circ$) and $p<0.01$ ($^*$).}
    \label{tab:results_auto}
\end{table}
Table~\ref{tab:results_auto} shows the results for automatic evaluation. We observe that ON/OFF outperforms ISO on MUST-C $D_1$, with respect to Sm(oothness) and F(luency) with relative improvements ranging from 9.9\%-28.3\% and 6.1\%-17.1\% respectively, while for In(telligibility) ON/OFF provides 0.6\%-14.5\% improvements for it, de, es. Similar improvements are obtained for ON/OFF against ISO for MuST-C on $D_2$ and YouTube on $D_1$ and $D_2$.  
We note that improvements for $D_1$ are higher compared to $D_2$, primarily because $D_1$ considers videos in which all sentences are off-screen. This provides more opportunities for ON/OFF to exploit the global relaxation mechanism. 

Compared to MuST-C, we find that YouTube data obtains higher In scores across all target languages and datasets. To investigate this further, we computed the length compliance (LC) metric of \cite{lakew2021isometricmt} at the phrasal level that measures the percentage of translations whose length in character is within $\pm 10\%$ of the length of the source and hence more suitable for auto dubbing. We found that on average, LC for YouTube was 19.8\% higher than that for MuST-C. Higher value of LC implies that we are able to better fit the translations in the available phrase intervals thereby having TTS speech more intelligible.

\subsection{Human Evaluation}   
\begin{table}[t]
    \centering
    \setlength\tabcolsep{5pt}
    \begin{tabular}{l|l|l|ll|ll}
      & &    & \multicolumn{2}{c|}{$D_1$} &    \multicolumn{2}{c}{$D_2$} \\
      & &    & \multicolumn{2}{l|}{ISO\hspace{0.1cm}    \hspace{0.0cm} ON/OFF} & \multicolumn{2}{l}{ISO\hspace{0.1cm}     \hspace{0cm} ON/OFF} \\
      \hline
      %Wins  \hspace{0cm} & 28.37\% & 46.33\% ** & 29.9\% & 42.24\% ** \\
    \rottext{8}{MuST-C}& fr & W  \hspace{0cm} & 18.3 & 38$^*$ & 21 & 43$^*$ \\
     && S  & 4.43 & 4.79$^*$   & 4.35 & 4.71$^*$ \\
    & it & W  \hspace{0cm} & 23 & 53.7$^*$ & 15.3 & 54.3$^*$\\
     && S  & 4.61 & 5.36$^*$ & 4.87 & 5.64$^*$ \\
    & de & W  \hspace{0cm} & 23.3 & 55.7$^*$ & 19.7 & 64.7$^*$\\
     && S  & 4.45 & 5.11$^*$ & 3.92 & 5.04$^*$ \\
    & es & W  \hspace{0cm} & 16.7 & 36.7$^*$ & 28.7 & 37.3$^*$\\
     && S  & 5.03 & 5.35$^*$ & 5.21 & 5.3 \\
    \hline
    \rottext{8}{YouTube} & fr & W  \hspace{0cm} & 21.5 & 58.2 $^*$ & 20.0 & 60.0$^*$\\
    && S  & 5.06 & 5.77$^*$  & 4.68 &  5.35$^*$\\
    & it & W  \hspace{0cm} & 17.3 & 68.7$^*$ & 21.7 & 55.0$^*$\\
    && S  & 5.03 & 6.16$^*$  & 5.08 & 5.87$^*$ \\
    & de & W  \hspace{0cm} & 20.3 & 56.7$^*$ & 25.7 & 50.0$^*$\\
    && S  & 5.35 & 6.17$^*$  & 5.00 & 5.53$^*$ \\
    & es & W  \hspace{0cm} & 27.0 & 56.7$^*$ & 36.3 & 46.3$^*$\\
    && S  & 4.70 & 5.36$^*$  & 4.95 & 5.09 \\
    \hline
    \end{tabular}
    \caption{Manual evaluations using Wins (W) and Score (S) with prosodic alignments: (ISO)  previous work on Isochrone dubbing \cite{virkar_improvements_2021}, (ON/OFF) new PA model for dubbing applied on off-screen clips ($D_1$) or on mixed off-screen and on-screen clips ($D_2$). All test sets consist of 15 4-sentence video clips for each domain (MuST-C, YouTube). Significance testing is done with levels $p<0.05$ ($^\circ$) and $p<0.01$ ($^*$).}
    \label{tab:results}
\end{table}
In this section, we present results of human evaluation on the test set. For each dubbing direction and dataset we report results on 15 video clips extracted separately for each evaluation using the criterion noted in Sec.~\ref{sec:evaluation_data}.  We asked 20 native speakers to rate the subjective experience for viewing each dubbed video from each dubbing condition on a scale of 0-10. To reduce cognitive load, we compare two dubbing conditions for each evaluation and collect a total of 600 scores. For all dubbing conditions we utilize post-edited translations to focus the subjects on the synchronization aspect of dubbing. 

Finally, for each evaluation we compare two conditions in a head-to-head manner and report Wins (percentage of times one condition is preferred over the other) and Score (average subjective score of dubbed videos) metrics. To measure the impact of PA model on human score, we use a linear-mixed-effects model (LMEM) \footnote{We used the lme4 package for R \cite{bates_fitting_2015}} by defining subjects and clips as random effects \cite{bates_parsimonious_2015}. 

The results are summarized in Table~\ref{tab:results}. For the dubbing evaluations, we compare ON/OFF vs ISO on test sets $D_1$ and $D_2$ in two separate evaluations. ON/OFF clearly outperforms ISO on $D_1$ providing relative improvements in Wins on both MuST-C ranging from 107.7\%-139.1\% 
and YouTube ranging from 110\%-297.1\%
with all results being statistically significant ($p < 0.01$). Similarly, ON/OFF outperforms ISO on $D_2$. Finally for Score, we obtain similar relative improvements on both MuST-C (1.7\%-28.6\%) and YouTube (2.8\%-22.5\%) with all improvements except the ones for es on $D_2$ being statistically significant ($p < 0.01$). 
    
\begin{table}[t]
    \centering
    \setlength\tabcolsep{5pt}
    \begin{tabular}{ll|ll|ll}
              &        & \multicolumn{2}{c|}{MuST-C} & \multicolumn{2}{c}{YouTube} \\ 
              &        & $D_1$    & $D_2$ & $D_1$ & $D_2$ \\ \hline
     {MuST-C} & $D_1$  & 0.51$^*$ & 0.43$^*$     & 0.73$^*$ & 0.70$^*$\\
              & $D_2$  & 0.07     & 0.26$^\circ$ & 0.50$^*$ & 0.50$^*$\\ \hline

     {YouTube}& $D_1$  & 0.14     & 0.33$^\circ$ & 0.68$^*$ & 0.65$^*$ \\
              & $D_2$  & 0.43$^*$ & 0.47$^*$     & 0.68$^*$ & 0.7$^*$  \\ 
             
    \end{tabular}
    \caption{Pearson correlation coefficient between predicted score from a LMEM model with fixed effects Sm, F, In and the averaged human score. We train a LMEM model on dataset in each row and predict score on dataset in each column. Significance testing is done with levels $p<0.05$ ($^\circ$) and $p<0.01$ ($^*$).}
    \label{tab:corr}
\end{table}

{\bf Relation between automatic and human scores: }
To explain the observed score variations using automatic metrics, we utilize LMEMs by aggregating evaluation data across all four languages. We define automatic metrics as fixed effects and subjects, clips, PA models and target languages as random effects. Our analysis reveals that Sm is the most impactful metric which is always statistically significant. 

To further test if the learned LMEM can help predict human score, for every example we compute the average human score and compare it with the predicted score. We average out the random effect of subjects since different subjects use different range to grade their viewing experience. We train LMEM models, one on each dataset and domain, for a total of 4 models and predict scores on all 4 datasets. Table~\ref{tab:corr} shows that in most cases we obtain a high pearson correlation coefficient between the predicted and average human score. Additionally, each model is able to correctly predict on average the winning system (ISO or On/Off) for all datasets. In some cases, due to the low correlation, the magnitude of difference in predicted scores between the systems is not consistent with actual score differences. However, the sign of the difference is consistent and hence we can predict the correct winning system even in these cases. Thus we conclude that automatic metrics can also help compare dubbing quality of two systems.

\section{Conclusions}

We extended prosodic alignment to off-screen dubbing that requires less stringent synchronization constraints. We address off-screen dubbing by introducing a global relaxation algorithm in which we relax timing constraints across all off-screen sentences and compute optimal relaxations using dynamic programming. Both automatic and human evaluations show that compared to applying isochrone dubbing for all sentences, relaxing the synchronization constraints for off-screen sentences significantly improves model performance on both automatic and subjective metrics. Finally, the analysis using linear mixed-effects models shows that a linear combination of all automatic metrics correlates well with the average human score and can be useful to compare dubbing quality of two systems.

\bibliographystyle{IEEEbib.bst}
\bibliography{biblio}

\begin{thebibliography}{10}

\bibitem{casacuberta_recent_2008}
F.~Casacuberta, M.~Federico, H.~Ney, and E.~Vidal,
\newblock ``Recent efforts in spoken language translation,''
\newblock {\em IEEE Signal Processing Magazine}, vol. 25, no. 3, pp. 80--88,
  2008.

\bibitem{weiss_sequence--sequence_2017}
R.~J. Weiss, J.~Chorowski, N.~Jaitly, Y.~Wu, and Z.~Chen,
\newblock ``Sequence-to-{Sequence} {Models} {Can} {Directly} {Translate}
  {Foreign} {Speech},''
\newblock in {\em Proc. Interspeech}, 2017, pp. 2625--2629.

\bibitem{cross_vila_end--end_2018}
L.~C. Vila, C.~Escolano, J.~A.~R. Fonollosa, and M.~R. Costa-Jussà,
\newblock ``End-to-{End} {Speech} {Translation} with the {Transformer},''
\newblock in {\em {IberSPEECH}}, 2018, pp. 60--63.

\bibitem{sperber_speech_2020}
M.~Sperber and M.~Paulik,
\newblock ``Speech {Translation} and the {End}-to-{End} {Promise}: {Taking}
  {Stock} of {Where} {We} {Are},''
\newblock in {\em Proceedings of the 58th {Annual} {Meeting} of the
  {Association} for {Computational} {Linguistics}}, 2020, pp. 7409--7421.

\bibitem{chaume_synchronization_2004}
F.~Chaume,
\newblock ``Synchronization in dubbing: {A} translation approach,''
\newblock in {\em Topics in {Audiovisual} {Translation}}, pp. 35--52. 2004.

\bibitem{oktem_prosodic_2019}
A.~Öktem, M.~Farrús, and A.~Bonafonte,
\newblock ``Prosodic {Phrase} {Alignment} for {Machine} {Dubbing},''
\newblock in {\em Proceedings of {Interspeech}}, Graz, Austria, 2019,
\newblock arXiv: 1908.07226.

\bibitem{federico_speech--speech_2020}
M.~Federico, R.~Enyedi, R.~Barra-Chicote, R.~Giri, U.~Isik, A.~Krishnaswamy,
  and H.~Sawaf,
\newblock ``From {Speech}-to-{Speech} {Translation} to {Automatic} {Dubbing},''
\newblock in {\em Proceedings of the 17th {International} {Conference} on
  {Spoken} {Language} {Translation}}, Online, July 2020, pp. 257--264,
  Association for Computational Linguistics.

\bibitem{federico_evaluating_2020}
M.~Federico, Y.~Virkar, R.~Enyedi, and R.~Barra-Chicote,
\newblock ``Evaluating and optimizing prosodic alignment for automatic
  dubbing,''
\newblock in {\em Proceedings of {Interspeech}}, 2020, p.~5.

\bibitem{virkar_improvements_2021}
Y.~Virkar, M.~Federico, R.~Enyedi, and R.~Barra-Chicote,
\newblock ``Improvements to {Prosodic} {Alignment} for {Automatic} {Dubbing},''
\newblock in {\em {ICASSP} 2021 - 2021 {IEEE} {International} {Conference} on
  {Acoustics}, {Speech} and {Signal} {Processing} ({ICASSP})}, June 2021, pp.
  7543--7574,
\newblock ISSN: 2379-190X.

\bibitem{saboo_integration_2019}
A.~Saboo and T.~Baumann,
\newblock ``Integration of {Dubbing} {Constraints} into {Machine}
  {Translation},''
\newblock in {\em Proceedings of the {Fourth} {Conference} on {Machine}
  {Translation} ({Volume} 1: {Research} {Papers})}, Florence, Italy, Aug. 2019,
  pp. 94--101, Association for Computational Linguistics.

\bibitem{lakew_controlling_2019}
S.~M. Lakew, M.~Di~Gangi, and M.~Federico,
\newblock ``Controlling the {Output} {Length} of {Neural} {Machine}
  {Translation},''
\newblock in {\em Proceedings of {IWSLT}}, Hong Kong, China, Oct. 2019,
\newblock arXiv: 1910.10408.

\bibitem{MUSTC_2019}
Mattia~A. Di~Gangi, Roldano Cattoni, Luisa Bentivogli, Matteo Negri, and Marco
  Turchi,
\newblock ``{MuST}-{C}: a {Multilingual} {Speech} {Translation} {Corpus},''
\newblock in {\em Proc. NAACL}, 2019, pp. 2012--2017.

\bibitem{vaswani_attention_2017}
A.~Vaswani, N.~Shazeer, N.~Parmar, J.~Uszkoreit, L.~Jones, A.~N. Gomez,
  L.~Kaiser, and I.~Polosukhin,
\newblock ``Attention {Is} {All} {You} {Need},''
\newblock {\em arXiv:1706.03762 [cs]}, 2017,
\newblock arXiv: 1706.03762.

\bibitem{di_gangi_robust_2019}
M.~A. Di~Gangi, R.~Enyedi, A.~Brusadin, and M.~Federico,
\newblock ``Robust {Neural} {Machine} {Translation} for {Clean} and {Noisy}
  {Speech} {Transcripts},''
\newblock in {\em Proc. {IWSLT}}, 2019.

\bibitem{prateek_other_2019}
N.~Prateek, M.~Lajszczak, R.~Barra-Chicote, T.~Drugman, J.~Lorenzo-Trueba,
  T.~Merritt, S.~Ronanki, and T.~Wood,
\newblock ``In {Other} {News}: a {Bi}-style {Text}-to-speech {Model} for
  {Synthesizing} {Newscaster} {Voice} with {Limited} {Data},''
\newblock in {\em Proceedings of the 2019 {Conference} of the {North}
  {American} {Chapter} of the {Association} for {Computational} {Linguistics}:
  {Human} {Language} {Technologies}, {Volume} 2 ({Industry} {Papers})},
  Minneapolis, Minnesota, 2019, pp. 205--213, Association for Computational
  Linguistics.

\bibitem{latorre_effect_2018}
J.~Latorre, J.~Lachowicz, J.~Lorenzo-Trueba, T.~Merritt, T.~Drugman,
  S.~Ronanki, and K.~Viacheslav,
\newblock ``Effect of data reduction on sequence-to-sequence neural {TTS},''
\newblock {\em arXiv:1811.06315 [cs, eess]}, 2018,
\newblock arXiv: 1811.06315.

\bibitem{lorenzo-trueba_towards_nodate}
J.~Lorenzo-Trueba, T.~Drugman, J.~Latorre, T.~Merritt, B.~Putrycz,
  R.~Barra-Chicote, A.~Moinet, and V.~Aggarwal,
\newblock ``Towards {A}chieving {R}obust {U}niversal {N}eural {V}ocoding,''
\newblock in {\em Proc. {Interspeech}}, 2019.

\bibitem{ronneberger2015u}
O.~Ronneberger, P.~Fischer, and T.~Brox,
\newblock ``U-net: Convolutional networks for biomedical image segmentation,''
\newblock in {\em Proc. ICMAI}. Springer, 2015, pp. 234--241.

\bibitem{jansson_singing_2017}
A.~Jansson, E.~Humphrey, N.~Montecchio, R.~Bittner, A.~Kumar, and T.~Weyde,
\newblock ``Singing voice separation with deep {U}-net convolutional
  networks,''
\newblock in {\em Proceedings of the 18th {International} {Society} for {Music}
  {Information} {Retrieval} {Conference}}, Suzhou, China, 2017, p.~8.

\bibitem{lollmann2010improved}
H.~L{\"o}llmann, E.~Yilmaz, M.~Jeub, and P.~Vary,
\newblock ``An improved algorithm for blind reverberation time estimation,''
\newblock in {\em Proc. IWAENC}, 2010, pp. 1--4.

\bibitem{habets2006room}
E.~A.~P. Habets,
\newblock ``Room impulse response generator,''
\newblock Tech. {R}ep. 2.4, Technische Universiteit Eindhoven, 2006.

\bibitem{karakanta_two_2020}
A.~Karakanta, S.~Bhattacharya, S.~Nayak, T.~Baumann, M.~Negri, and M.~Turchi,
\newblock ``The {Two} {Shades} of {Dubbing} in {Neural} {Machine}
  {Translation},''
\newblock in {\em Proceedings of the 28th {International} {Conference} on
  {Computational} {Linguistics}}, Barcelona, Spain (Online), 2020, pp.
  4327--4333, International Committee on Computational Linguistics.

\bibitem{ochshorn_gentle_2017}
R.~M. Ochshorn and M.~Hawkins,
\newblock ``Gentle {Forced} {Aligner},'' 2017.

\bibitem{mcauliffe_montreal_2017}
M.~McAuliffe, M.~Socolof, S.~Mihuc, M.~Wagner, and M.~Sonderegger,
\newblock ``Montreal {Forced} {Aligner}: {Trainable} {Text}-{Speech}
  {Alignment} {Using} {Kaldi},''
\newblock in {\em Interspeech}, 2017, pp. 498--502.

\bibitem{di_gangi_must-c:_2019}
M.~A. Di~Gangi, R.~Cattoni, L.~Bentivogli, M.~Negri, and M.~Turchi,
\newblock ``{MuST}-{C}: a {Multilingual} {Speech} {Translation} {Corpus},''
\newblock in {\em Proceedings of the 2019 {Conference} of the {North}
  {American} {Chapter} of the {Association} for {Computational} {Linguistics}:
  {Human} {Language} {Technologies}, {Volume} 1 ({Long} and {Short} {Papers})},
  Minneapolis, Minnesota, 2019, pp. 2012--2017, Association for Computational
  Linguistics.

\bibitem{lakew2021isometricmt}
Surafel~M Lakew, Yogesh Virkar, Prashant Mathur, and Marcello Federico,
\newblock ``Isometric mt: Neural machine translation for automatic dubbing,''
\newblock {\em arXiv preprint arXiv:2112.08682}, 2021.

\bibitem{bates_fitting_2015}
Douglas Bates, Martin Mächler, Ben Bolker, and Steve Walker,
\newblock ``Fitting {Linear} {Mixed}-{Effects} {Models} {Using} lme4,''
\newblock {\em Journal of Statistical Software}, vol. 67, no. 1, pp. 1--48,
  Oct. 2015.

\bibitem{bates_parsimonious_2015}
Douglas Bates, Reinhold Kliegl, Shravan Vasishth, and Harald Baayen,
\newblock ``Parsimonious {Mixed} {Models},''
\newblock {\em arXiv:1506.04967 [stat]}, June 2015,
\newblock arXiv: 1506.04967.

\end{thebibliography}
\end{document}